\pdfoutput=1

\documentclass[11pt]{article}

\usepackage[final]{acl}
\usepackage{listings}
\usepackage{xcolor}
\usepackage{times}
\usepackage{latexsym}
\usepackage{multirow}
\usepackage{multicol}
\usepackage{amsmath}
\usepackage{mathrsfs}
\usepackage[T1]{fontenc}

\usepackage[utf8]{inputenc}

\usepackage{microtype}

\usepackage{inconsolata}

\usepackage{graphicx}

\colorlet{punct}{red!60!black}
\definecolor{background}{HTML}{EEEEEE}
\definecolor{delim}{RGB}{20,105,176}
\colorlet{numb}{magenta!60!black}
\lstdefinelanguage{json}{
    basicstyle=\normalfont\ttfamily,
    numbers=left,
    numberstyle=\scriptsize,
    stepnumber=1,
    numbersep=8pt,
    showstringspaces=false,
    breaklines=true,
    frame=lines,
    backgroundcolor=\color{background},
    literate=
     *{0}{{{\color{numb}0}}}{1}
      {1}{{{\color{numb}1}}}{1}
      {2}{{{\color{numb}2}}}{1}
      {3}{{{\color{numb}3}}}{1}
      {4}{{{\color{numb}4}}}{1}
      {5}{{{\color{numb}5}}}{1}
      {6}{{{\color{numb}6}}}{1}
      {7}{{{\color{numb}7}}}{1}
      {8}{{{\color{numb}8}}}{1}
      {9}{{{\color{numb}9}}}{1}
      {:}{{{\color{punct}{:}}}}{1}
      {,}{{{\color{punct}{,}}}}{1}
      {\{}{{{\color{delim}{\{}}}}{1}
      {\}}{{{\color{delim}{\}}}}}{1}
      {[}{{{\color{delim}{[}}}}{1}
      {]}{{{\color{delim}{]}}}}{1},
}

\title{Aggregated Structural Representation with Large Language Models for Human-Centric Layout Generation}

\author{
  \textbf{Jiongchao Jin\textsuperscript{1,2}},
  \textbf{Shengchu Zhao\textsuperscript{1}},
  \textbf{Dajun Chen\textsuperscript{1}},
  \textbf{Wei Jiang\textsuperscript{1}},
  \textbf{Yong Li\textsuperscript{1}}
\\
\\
  \textsuperscript{1}Ant Group,
  \textsuperscript{2}Agency for Science, Technology and Research
\\
  \small{
    \textbf{Correspondence:} \href{mailto:email@domain}{jin\_jiongchao@ihpc.a-star.edu.sg}
  }
}

\makeatletter
\AtBeginDocument{
\let\@oldmaketitle\@maketitle
\renewcommand{\@maketitle}{\@oldmaketitle
  \vspace{-27pt}
  \includegraphics[width=\linewidth]{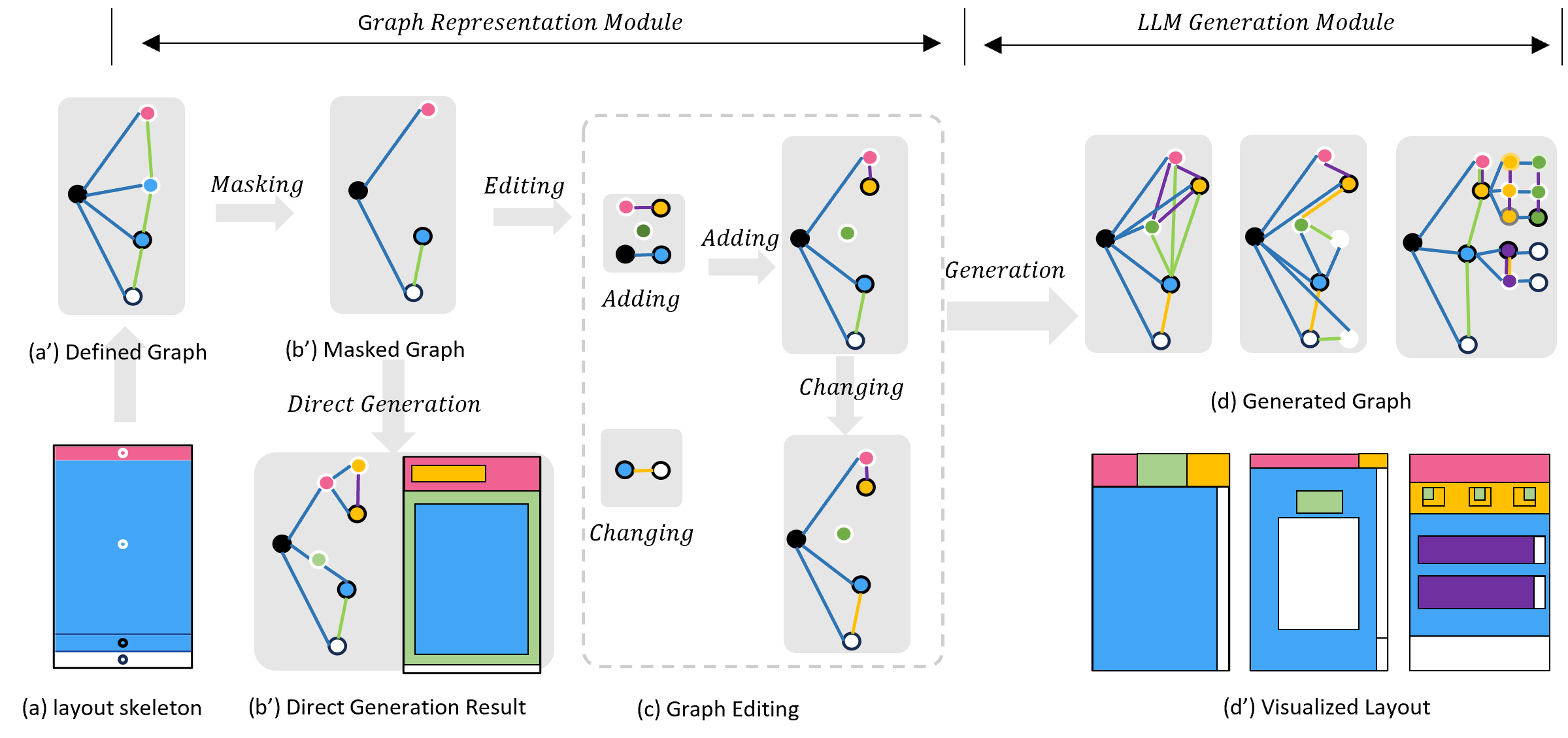}
  \vspace{-18pt}
  \captionof{figure}{
    ASR extracts hierarchical representations from graphs with semantic and positional relationships, then aggregates them with a large language model for layout generation. Human interaction during the process enables customization, resulting in diverse, user-centered layouts adaptable to practical design needs.
  }
  \label{fig:teaser}
  \vspace{10pt}
 }
}
\makeatother

\begin{document}
\maketitle

\begin{abstract}
 \vspace{-0.5em}
Time consumption and the complexity of manual layout design make automated layout generation a critical task, especially for multiple applications across different mobile devices. Existing graph-based layout generation approaches suffer from limited generative capability, often resulting in unreasonable and incompatible outputs. Meanwhile, vision-based generative models tend to overlook the original structural information, leading to component intersections and overlaps. To address these challenges, we propose an Aggregation Structural Representation (ASR) module that integrates graph networks with large language models (LLMs) to preserve structural information while enhancing generative capability. This novel pipeline utilizes graph features as hierarchical prior knowledge, replacing the traditional Vision Transformer (ViT) module in multimodal large language models(MLLM) to predict full layout information for the first time. Moreover, the intermediate graph matrix used as input for the LLM is human-editable, enabling progressive, human-centric design generation. A comprehensive evaluation on the RICO dataset demonstrates the strong performance of ASR, both quantitatively—using mean Intersection over Union (mIoU), —and qualitatively through a crowd-sourced user study. Additionally, sampling on relational features ensures diverse layout generation, further enhancing the adaptability and creativity of the proposed approach.
\end{abstract}

\section{Introduction}

The growing demand for various terminal devices has led to increasingly diverse user interfaces. However, manually designing user interface(UI) layouts is a time-consuming process. Moreover, with the rapid development of large language models and generative models, an abundance of UI layouts and images is required for UI-based tasks such as UI understanding and terminal-based agents. Consequently, layout generation using generative AI remains a challenging topic in structural and visual representation learning.

Traditional layout generation methods rely on graph-based or structural learning approaches. However, due to the varied and uncertain nature of UI images, these methods suffer from artifacts such as overlaps, incompatibility, and unreasonable layout designs. Current MLLM-based layout generation techniques alleviate issues but tend to overlook the original structural information. 

In this paper, we propose a novel aggregating pipeline, Aggregation Structural Representation (ASR), which combines the graph hierarchical structure with a large language model. This approach preserves the structural information, making it available for human-centered graph editing, while enhancing the model's generative ability, resulting in more accurate layout generation.

Accurate structural information and bounding box coordinates are crucial for resolving overlaps in UI layout generation. ASR addresses this by defining semantic (e.g., contain, parallel) and positional (e.g., left, top) relationships between bounding boxes to avoid meaningless component intersections. The pipeline combines a graph neural network for extracting structural representations with a decoder that supervises component relationships, ensuring a guided and accurate generation process for the large language model (LLM). Multimodal LLM (M-LLMs) offer superior prediction and generation capabilities compared to traditional graph networks. To harness this potential, we use the LLM to recover and generate fundamental component attributes. Its grounding capability enables precise bounding box coordination, while its classification ability accurately predicts component types and other attributes, resulting in coherent and diverse layout generation.

The ASR pipeline emphasizes human-centric generation by enabling users to actively influence the design process. A key component is the editable relation matrix, which users can modify manually or through randomly generated factors to adjust graph relations and guide the layout generation. This interaction allows for personalized, diverse layouts in various formats and styles, ensuring that the generated designs meet specific user needs and preferences. Our adopted M-LLM processes both graph features and the relation matrix simultaneously, with the relation matrix acting as a supervisory signal to maintain structural consistency. Even small changes to the relation matrix can lead to significant transformations in the final layout, empowering users to explore a broad range of design possibilities. The results section highlights this adaptability, demonstrating the flexibility and human-centered nature of our method.

\section{Related Works}

\subsection{Graph Neural Network}
Graph neural networks (GNNs) are well-suited for extracting graph features that include both node and relational information, making them effective for handling scenarios with structural and hierarchical data, such as scene generation \cite{gao2020multi}, floorplan understanding \cite{graphist2023hlg}, and layout understanding \cite{ren2020paragraph}. 

In the context of UI layout generation, Manandhar et al. \cite{manandhar2020learning} employed a GCN-CNN architecture on a graph of UI layout images, trained under an IoU-based triplet network \cite{hoffer2015deep}. However, their approach computes graph embeddings for the anchor, positive, and negative graphs independently. Subsequently, LayoutGMN \cite{patil2021layoutgmn} advanced this by learning graph embeddings in a dependent manner, utilizing cross-graph information exchange to learn embeddings in the context of anchor-positive and anchor-negative pairs. Unlike the GCN-CNN framework, which trains triplet networks using IoUs, LayoutGMN argued that IoU does not account for structural matching, making it an unreliable metric for structural similarity.

In contrast, our proposed Aggregating Structural Representation (ASR) aims to enhance structural representation aggregation by combining GNNs with a large language model (LLM) in the generation phase. While we still use GNN for graph representation learning, we address the limitations of traditional GNN-based pipelines by incorporating LLMs, which work under the supervision of the GNN features and relation matrix, improving the overall layout generation process.

\begin{table*}[t]
    \centering
    \begin{tabular*}{\linewidth}{cc|cc|ccc}
    \hline
        {Param} & Meaning & Func & Meaning & Graph Def &  Type& Value\\  
        \hline
        
        $N$ & Nodes & \multirow{2}*{$MLP$} & Multilayer- & \multirow{8}*{Attributes} &\multirow{6}*{Category}& BACKGROUND  \\
        \cline{1-2} \cline{7-7}
        $E$ &Edges&&Perception & & &IMAGE \\
        \cline{1-4} \cline{7-7}
        $\mathcal{G}$ &Graph & $\mathscr{F}$ & Graph Encoder && &TEXT \\
         \cline{1-4} \cline{7-7}
        $N_f$ & Node Feature & $concat$ & Concat Layer &&& SLIDING BAR \\
        \cline{1-4} \cline{7-7}
        $text$ & Text & \multirow{2}* {$ViT$} & Vision- &&& ICON\\
        \cline{1-2} \cline{7-7}
        $coord$&coordinate&&Transformer &&& INPUT\\
        \hline
        $I_{icon}$ & Icon image &$Tk$& Tokenizer &  &Coordinate& [X,Y,W,H] \\
        \cline{1-4}\cline{6-7}
        \multirow{2}*{$I_{image}$} & Component- & $\mathscr{D}$ & Graph Decoder &\multirow{4}*{Relations}&\multirow{2}*{Positional}&TOP\\
        \cline{3-4} \cline{7-7}
        & Image && &&&LEFT\\
        \cline{1-2} \cline{6-7}
        $h_a$& Graph Feature && &&\multirow{2}*{Semantic}&PARALLEL\\
        \cline{1-2}\cline{7-7}
        \multirow{2}*{$G_{gt}$} & Ground Truth  &&&&&\\
        &Graph Matrix & & &&&CONTAIN \\
        \hline
    \end{tabular*}
    \caption{Defination of Parameters, Functions and Graph Details}
    \vspace{-1em}
    \label{tab:params}
\end{table*}

\subsection{Large Language Model}
Recent advancements in large language models (LLMs) have led to significant progress in various fields. Models such as GPT \cite{OpenAI2023GPT-4}, LLaMA \cite{LLaMa1,LLaMa2,LLaMa3}, and Qwen \cite{Qwen2023,Qwen2024} have demonstrated impressive capabilities across a wide range of applications. Different domains, including embodied artificial intelligence \cite{driess2023palme,2023embodiedgpt,rt22023arxiv}, mobile phone operation agents \cite{wang2024mobile,yang2023appagent,nong2024mobileflow}, and layout generation \cite{lin2023layoutprompter,seol2024posterllama,sasazawa2024layout}, have been profoundly influenced by LLMs.

We adopt multimodal large language models (MLLMs), which extend traditional LLMs by integrating non-text modalities (e.g., images, videos, graphs) through an encoder, base model, and connector. MLLMs bridge text and other data, enabling more comprehensive understanding and generation. Widely applied in fields like mobile agents \cite{hong2023cogagent,wang2025mobileagenteselfevolvingmobileassistant,wang2024mobile2}, autonomous driving \cite{yuan2024rag,wei2024editable}, image generation \cite{qin2024diffusiongpt,wang2024genartist}, video understanding \cite{tang2023video,li2023videochat}, and other fields. MLLMs inspire our novel graph-and-LLM-based layout generation method, combining graph information with LLM’s generative capabilities.

\subsection{Layout Generation}
The layout generation is an important part of the generation tasks. A large number of application scenarios, such as UI design\cite{duan2024efficient,rahman2021ruite}, poster creation\cite{seol2024posterllama,yang2024posterllava,graphist2023hlg}, document layout design\cite{zhu2024automatic,he2023diffusion,yamaguchi2021canvasvae}, floor plan generation\cite{hu2020graph2plan,weber2022automated}, are designed to meet diverse user requirements. Layout generation methods can be divided into constraint-based methods\cite{bauerly2006computational} and generative model-based methods\cite{hsu2023densitylayout,li2023relation}. \citeauthor{yang2016automatic} present a constraint-based layout generation framework integrating aesthetic principles and image features to optimize graphic design. However, its reliance on predefined constraints limits generative diversity. \citeauthor{li2019layoutgan} introduce LayoutGAN, featuring a differentiable wireframe rendering layer and CNN-based discriminator to capture layout styles and element relationships. \citeauthor{Jiang_2023_CVPR} propose LayoutFormer++, a conditional graphic layout generation model that offers flexibility and controllability through constraint serialization and decoding space restriction strategies. More recently, \citeauthor{lin2023layoutprompter} illustrate a conditional graphic layout approach using large language models (LLMs) with in-context learning, combining input-output serialization, dynamic exemplar selection, and layout ranking for versatility and data efficiency. \citeauthor{graphist2023hlg} present Graphist, a layout generation model based on large multimodal models, introducing the Hierarchical Layout Generation (HLG) task to create graphic compositions from unordered design elements.

Inspired by these works, we propose a novel method that integrates multimodal LLMs with graphic relations to balance generative diversity and fundamental constraints.
\begin{figure*}[h]
    \centering
    \includegraphics[width=1.0\linewidth]{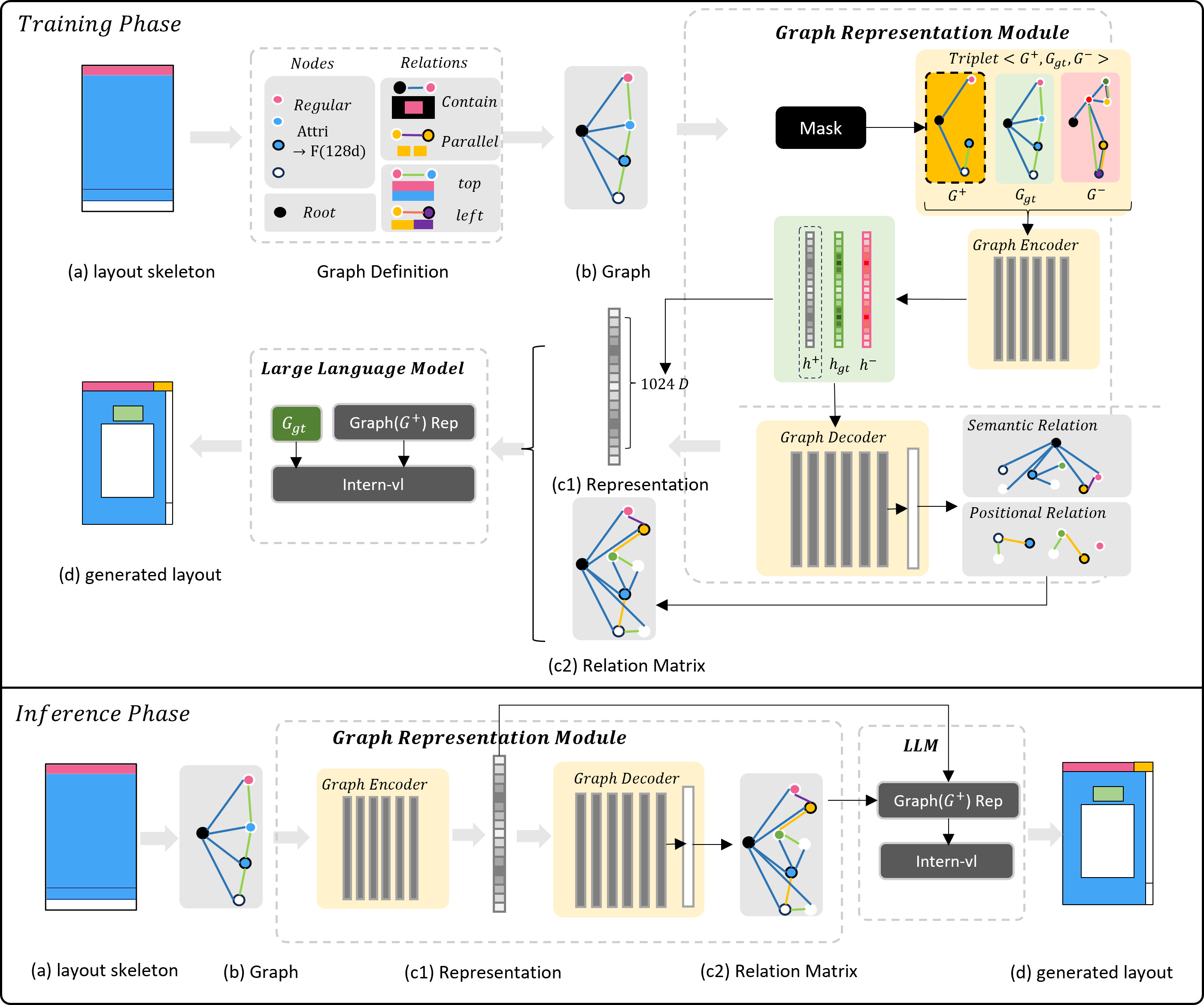}
    \caption{Overview pipeline of ASR. The top subfigure shows the training phase of ASR while the bottom subfigure indicates the inference phase of ASR. }
    \label{fig:Overview}
\end{figure*}

\section{Methods}
The pipeline overview and parameter definitions are provided in Figure \ref{fig:Overview} and Table \ref{tab:params}. In this section, we detail the modules of our proposed pipeline.

\subsection{Graph Definition $\mathcal{G}_{<N, E>}$}

Layout generation tasks can be formulated as graph $\mathcal{G}$ generation by defining nodes ($N$, representing components with attributes) and edges ($E$, representing relationships).
 
As shown in Table \ref{tab:params}, each graph node has two attributes: its category, indicating the component type, and its coordinate, specifying the component’s central position and size. In our definition, component categories include background, image, text, input (box or bar), sliding bar, and icon.

\begin{figure}[t]
\vspace{-1em}
    \centering
    \includegraphics[width=1\linewidth]{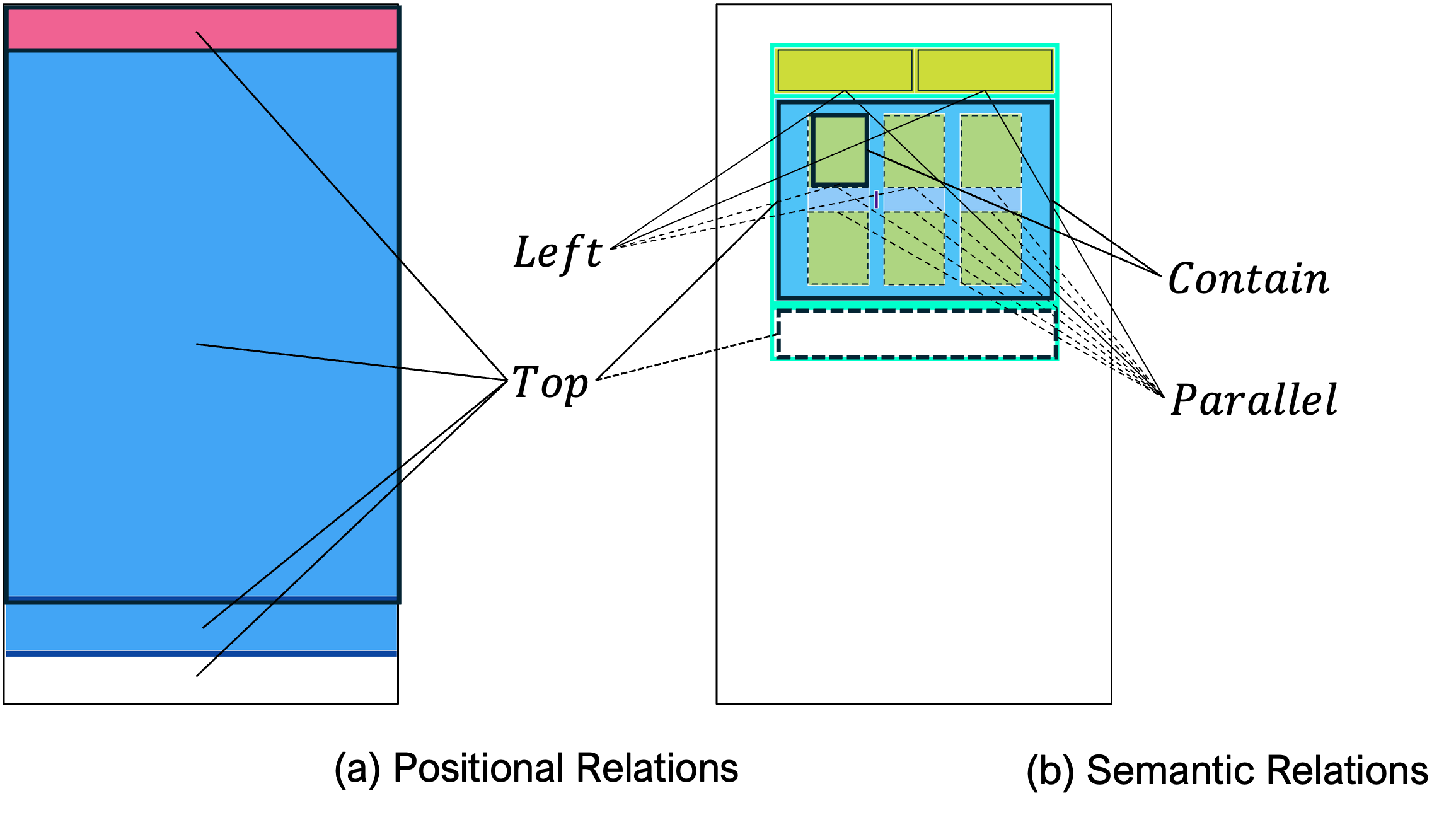}
    \caption{The explanation of semantic relations and positional relations}
    \vspace{-1em}
    \label{fig:Relations}
\end{figure}

We define positional relations to indicate whether one component (node) is above or to the left of another, and semantic relations to specify containment or parallel structures. A contain relation signifies that a component is nested within another—i.e., the contained component is a child node of the background (father) node. In contrast, a parallel relation indicates that components share the same hierarchy, each acting as a child node of the same father node. A visualization of these component relations is provided in Figure \ref{fig:Relations}

\subsection{Graph Representation Module $\mathscr{F}_\mathcal{G}$}
The graph representation module, shown in Figure \ref{fig:Overview}, employs a five-layer graph neural network (GNN) encoder and decoder during both training and inference. Additionally, mask and triplet generation are used only in the training phase.

First, after defining graph nodes and relations, we convert discrete node attributes into continuous features. Icons, images, and text are encoded using a Vision Transformer ($ViT$) and tokenizer, then combined with their coordinates and type information to form a 128-dimensional feature vector $N_f$, as shown in Equation \ref{e:2}.

\begin{equation}
\begin{split}
    \label{e:2}
N_f  &= MLP(ViT(I_{icon}, I_{image}), \\&Tk(text), coord, cat) 
\end{split}
\end{equation}

Then, each node’s feature $N_f$ is concatenated with the corresponding entries from the relation matrix $E_M$ along the node dimension $n$. Five layers of the GNN $\mathscr{F}_\mathcal{G}$ are then applied to extract the final 1024-dimensional feature $h_a$, as demonstrated in Equation \ref{e:1}.

\begin{equation}
\label{e:1}
h_a = \mathscr{F}_\mathcal{G}(concat(N_f, E_M)) \\
\end{equation}

In learning graph representation, we first re-arrange the $RICO$ dataset into triplets $\mathcal{T}\{G_{gt}, G^{+}, G^{-}\}$, where $G_{gt}, G^{+}, G^{-}$ represent the ground truth, positive, and negative layouts, respectively. In our experiments, $G^+$ is created by randomly masking 5\% to 25\% of the ground truth graph while $G^-$ is selected as an entirely different graph from the dataset.

During training, we employ the SimSCE\cite{gao2021simcse} loss to minimize the distance between $G_{p}$ and $G_{gt}$, while maximizing the distance between $G^{-}$ and $G_{gt}$.
The corresponding equations are shown below.
\begin{equation}
\begin{split}
    \mathcal{L} &=\\ &-\log \frac{\exp(\text{sim}(\mathscr{F}_\mathcal{G}(G_{gt}), \mathscr{F}_\mathcal{G}(G^+)) / \tau)}{\sum_{G^- \in \mathcal{N}} \exp(\text{sim}(\mathscr{F}_\mathcal{G}(G_{gt}), \mathscr{F}_\mathcal{G}(G^-)) / \tau)}
    \end{split}
\end{equation}
where $\mathcal{N}$ is the whole set of Negative samples, $\tau$ is the temperature parameter, $sim$ stands for similarity and we adopted cosine similarity distance.

\subsection{Relation Matrix $M$}
Based on comprehensive observations, we found that GNNs have a strong ability to retrieve the relation matrix, while LLMs are more capable of recovering attributes. Therefore, we first employ a GNN decoder to decode the relation feature $M_{sem}$ and $M_{pos}$ by equation \ref{eq:3}.

\begin{equation}
\label{eq:3}
    M_{sem}, M_{pos} = \mathscr{D}_\mathcal{G}(h_a)\\
\end{equation}

We also emphasize that the relation matrix is editable, either through randomly generated factors or human-centered graph editing, thereby ensuring both diversity and user-centered design in the overall generation process. After that, in our main pipeline, the relation matrix $M_{sem}$, $M_{pos}$, and representation $h_a$ are combined together to a 1024-dimension input feature using an MLP.

When training the Graph Representation Module with triplets, the positive graph $G^+$ is a masked version of $G_{gt}$. Since the relation matrix decoder \begin{lstlisting}[language=json,firstnumber=1,label={Code},caption={LLM Output Format},captionpos=b]
{
  "Node_id": "1",
  "Category": "SLIDING BAR",
  "Coordinate":[1, 33, 172, 208],
  "Top": [3,5,6,7,9,10,11],
  "Left":[2,3,4],
  "Parallel":[7,9],
  "Contain":[]
  }
}
{
    "Node_id": "2", ....
}
\end{lstlisting}
is supervised by $G_{gt}$, the generated relation matrix inherits more detailed information from $G_{gt}$ rather than being limited to the skeleton of $G^{+}$. Hence, the decoding of the relation matrix already gained part of the generative capability. Thus, during inference phase, the graph representation module can complete portions of the original input graph with the assistance of the generative capability when decoding relation matrix. 

\subsection{Large Language Model Aggregation}
In our pipeline, the MLLM receives the relation matrix \(\{M_{sem}, M_{pos}\}\) and the graph feature \(h_a\) as input. Traditionally, MLLMs concatenate image features extracted by a ViT with prompts and other settings. In contrast, the ASR generation branch replaces these image features with our graph representation to ensure structural information is preserved in the generated layouts. The input is computed as:

\begin{equation}
input = MLP(M_{sem}, M_{pos}, h_a)
\end{equation}
We then fine-tuned the Intern-VL model by replacing its ViT component with our graph representation learning module. 
Consequently, the output is provided in JSON format, containing relation information, node categories, and coordinates. The output format is shown in Listing \ref{Code}.

In fine-tuning, we use the feature \(h_a\) from the masked graph \(G^+\) while retaining the relation matrix from \(G_{gt}\) as a skeleton to bolster the LLM’s generative ability. Consequently, the LLM must not only recover node attributes already encoded in \(h_a\) but also predict newly introduced attributes dictated by the relation matrix, all under supervision from \(G_{gt}\). Additionally, the LLM can generate random nodes and integrate them into the layout without violating the existing relation matrix.
During real-time inference—regardless of whether the relation matrix is directly output by the decoder, modified by random factors, or edited by users—the LLM can accurately predict every node attribute in the relation matrix and introduce additional random nodes that preserve structural consistency.

\section{Experiments}
\subsection{Experiment Settings}
We adopted a filtered RICO dataset containing 68,475 UI structures, PubLayNet \cite{zhong2019publaynet} with over 360K machine-annotated document images (following the cleaning process in \cite{kikuchi2021constrained,lee2020neural,li2019layoutgan}), and the Magazine dataset \cite{zheng-sig19}, which contains more than 4,000 images, to evaluate the generalization ability of our models. Note that we trained our ASR separately on each dataset, dividing them into training, validation, and testing sets in a 7:2:1 ratio.

Training (for RICO dataset) graph representation learning phase (Phase 1) ultized 1 Nvidia-A100 GPU for 50 epochs(7-hours) and LLM training adopted 2 Nvidia-A100 GPUs for 3 epochs(21-hours). Phase 1 learning rate is $10^{-4}$ for 10 epochs and $10^{-5}$ for the rest epochs. Phase 2 learning rate is $10^{-6}$ for all epochs.
\subsection{Metrics and subtask settings}
For the evaluation of layout generation, we used m-IoU, FID and relation error for quantitative evaluation, and using perceptual user study for qualitative evaluation.

\textbf{Relation Error(RE) $\downarrow$} measures the correctness of the relations by calculating the Mean Squared Error (MSE) distance between the relation matrix and the ground truth. For those without relation matrix, we extract the relation graph and calculate the MSE between ground truth. 
\textbf{Maximum Intersection-over-Union (mIoU) $\uparrow$} measures the similarity between the bounding boxes of the same category label in the generated layouts and the ground-truth layouts.
\textbf{Frechet Inception Distance (FID) $\downarrow$}
measures the distributional distance between the feature representations of generated layouts and their corresponding ground truth. To calculate FID, a model is trained to classify whether an input layout is corrupted or not. 
\textbf{Overlap (OL) $\downarrow$}  measures the degree of overlap between each element pair within generated layouts. A well-designed layout typically exhibits fewer overlaps, as this indicates better spatial arrangement and component distribution. 

For layout generation, we evaluate our generation ability on direct UI generation task, UI completion task(with nodes and edge constraint), and Graph Editing Ability(Involving User interaction). The UI generation and completion task's setting are aligned with \cite{hui2023unifying}. Note that our ASR allows for manual graph editing during the generation process, whereas other models require all settings to be predefined. Therefore, when evaluating graph editing ability, we coarsely set some nodes and relations before generation for the other models (which is different from Completion, where we provide precise relations and nodes upfront and only need new ones to be generated). This evaluation examines whether the models can generate the correct components and relations based on the given instructions and whether the generated components are compatible with both existing and newly generated ones. 

\begin{table*}[]\small
    \centering
    \renewcommand\arraystretch{1.5}
    \begin{tabular}{cccccccccccccc}
    \hline
    
         \multirow{2}*{Task}&\multirow{2}*{Models}&\multicolumn{4}{c}{Rico} & \multicolumn{4}{c}{PublayNet}  & \multicolumn{4}{c}{Magazines} \\
         \cline{3-6}
         \cline{7-10}
         \cline{11-14}
         &&RE$\downarrow$&mIoU$\uparrow$&OL$\downarrow$ &FID$\downarrow$ & RE$\downarrow$ & mIoU$\uparrow$ & OL$\downarrow$ & FID$\downarrow$   & RE$\downarrow$ & mIoU$\uparrow$ & OL$\downarrow$& FID$\downarrow$  \\
         \hline
         \multirow{5}*{UI-Gen}&L-GMN&0.36 &0.44&57.33 &38.52&\textbf{0.43}& 0.36&32.08&41.30&0.42 & 0.28 & 53.54 & 48.97 \\
         &UniLayout&0.42&0.47&59.26&25.45&0.55& 0.45&21.33&33.05& 0.55 & 0.31 & 52.86 & 40.33\\
         &L-DM &0.48&0.51&57.62&23.28&0.53&0.42&19.86&\textbf{21.38}&0.52 & \textbf{0.39} & 48.42 & \textbf{32.71} \\ 
         &LDGM &0.49&0.62&55.31&24.43&0.51&0.46&\textbf{19.71}&27.32&0.51 & 0.36 & 49.38 & 37.42 \\
        & ASR &\textbf{0.27}&\textbf{0.63}&\textbf{52.43}&\textbf{23.08}&0.44&\textbf{0.50}&19.77&24.08&  \textbf{0.38} & 0.36 & \textbf{44.34} & 33.85\\
         \hline 
         &L-GMN& 0.28&0.57&54.75 &32.06&\textbf{0.36}&0.42&28.54&33.42& 0.35 & 0.32 & 50.93 & 40.22\\
         comp-&UniLayout&0.33&0.66&55.99&25.18&0.48& 0.47 &20.76&30.24& 0.50 & 0.34 & 48.05 & 31.94\\
         -letion&L-DM &0.35&0.68&52.31&17.79&0.44&0.47&18.40&\textbf{20.55}& 0.46 & 0.40 & 42.03 & 27.60\\ 
         &LDGM &0.26&0.66&54.08&\textbf{16.81}&0.46&0.45&18.24&24.62& 0.45 & 0.38 & \textbf{39.88} & 24.43 \\
        & ASR &\textbf{0.24}&\textbf{0.70}&\textbf{51.16}&17.21&\textbf{0.36}&\textbf{0.51}&\textbf{17.32}&21.77& \textbf{0.33} & \textbf{0.43} & 42.50 & \textbf{21.69} \\
         \hline 
         &L-GMN& 0.12 & 0.58&45.13 &18.34&0.20&0.49&11.37&18.08&0.15 & 0.46 & 12.48 & 17.79\\
         Graph&UniLayout& 0.14&0.62&31.72&11.67&0.25&0.58 &9.72&14.32& 0.17 & 0.49 & 11.02 & 12.70  \\
         Editing&L-DM &0.15&\textbf{0.69}&26.98&\textbf{7.92}&0.21&0.63&\textbf{7.33}&9.25& \textbf{0.09} & 0.51 & 9.83 & \textbf{8.54} \\ 
         &LDGM &0.12&0.65&27.36&9.08&0.24&\textbf{0.64}&7.48&11.37& 0.13 & \textbf{0.54} & 10.33 & 9.29  \\
        & ASR &\textbf{0.09}&0.68&\textbf{26.30}&8.45&\textbf{0.18}&\textbf{0.64}&7.58&\textbf{8.48}&\textbf{0.09} & 0.53 & \textbf{9.77} & 10.35\\
         \hline 
    \end{tabular}
    \caption{Quantitative Results on Rico Dataset, comparing ASR with other SOTAs including graph neural network based Layout GMN, visual based UniLayout, LayoutDM, LDGM.}
    \label{tab:Quantitative}
\end{table*}

\subsection{Quantitative Results}

In this section, we show the quantitative results of our ASR pipeline with other SOTA networks, including LayoutGMN\cite{patil2021layoutgmn}, UniLayout\cite{jiang2022unilayout}, LayoutDM \cite{inoue2023layoutdm} and LDGM\cite{hui2023unifying} on direct UI generation(U-GEN), UI completion, and graph editing ability. 

Table \ref{tab:Quantitative} shows the quantitative results of all three tasks under Max IoU, Relation Error, Overlap and FID metrics. The ASR model stands out across all three tasks (UI Generation, Completion, and Graph Editing). It consistently outperforms other models in terms of mIoU, RE, and OL, demonstrating its effectiveness in generating layouts with high structural fidelity, accuracy, and minimal overlap. The low FID values further emphasize that ASR produces more realistic and coherent layout components, making it an efficient and powerful tool for layout generation and graph editing.

\subsection{Perceptual Results}
\begin{figure*}[t]
    \centering
    \includegraphics[width=1\linewidth]{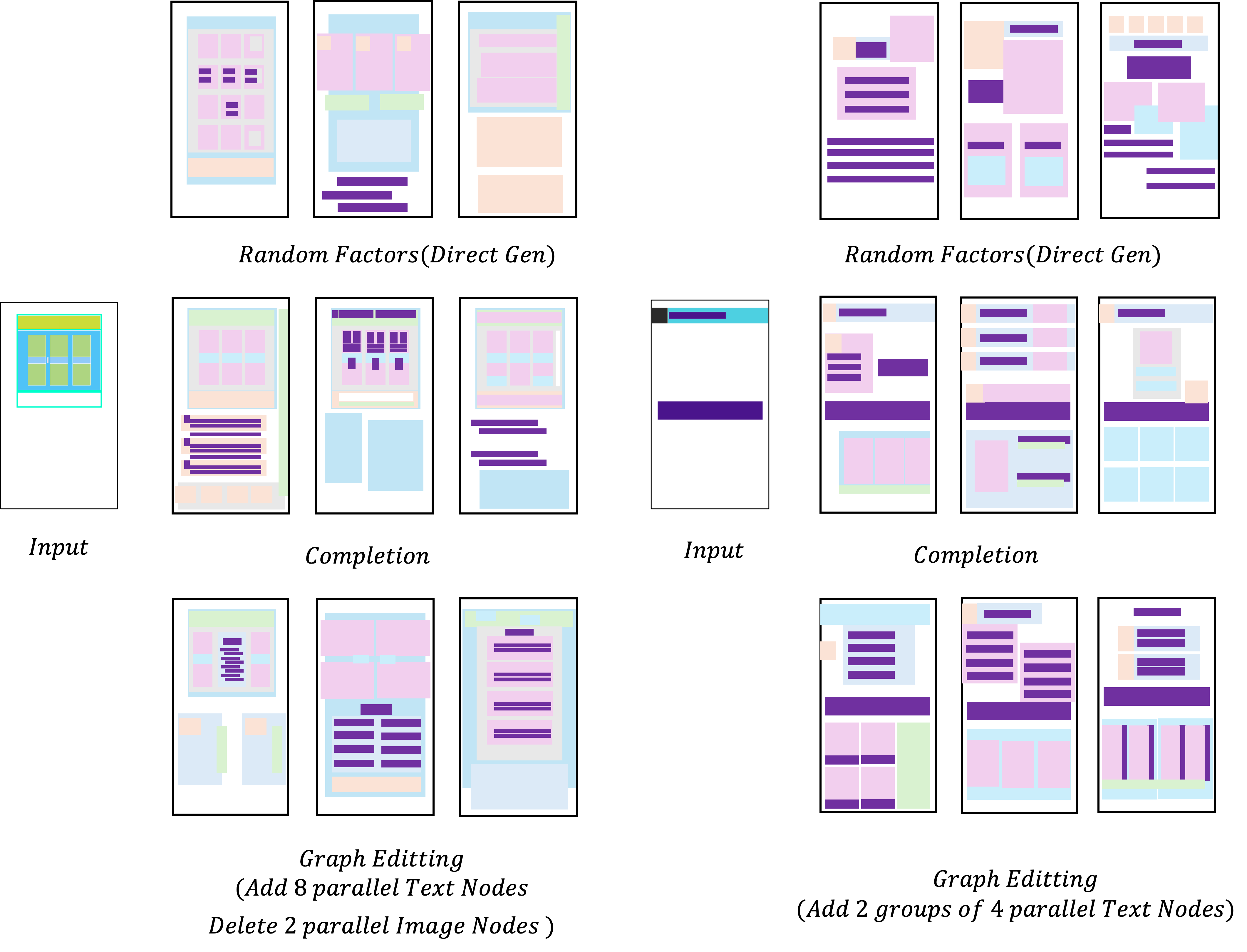}
    \caption{The gallery of qualitative results in different tasks}
    \label{fig:gallery}
\end{figure*}

To better leverage the perceptual of our generated layout designs, we did a crowd-source user study involving 50 people in Amanze Mechanical Turk.
In this perceptual study, after providing people with 5 generated layout designs together with original input, they are required to score the layouts from 0 to 5, indicating the number of reasonable layout in their mind. To better evaluate the generation ability, we divided all generated layouts into 3 different levels(easy, medium and hard) according to the number of components in one layout. Here, we define EASY when the number of components is less than 8, Medium means the number of components is between 9 and 20, Hard is more than 20. As shown in table, our ASR can generate more human friendly and practical layout results among easy, medium and hard type than any other SOTAs especially in hard layout samples. Besides, we also visualized some samples that our ASR on different tasks in Figure \ref{fig:gallery}

\begin{table}[]
    \centering
    \begin{tabular}{cccc}
    \hline
       &Easy&Medium&Hard\\ \hline
         LayoutGMN& 3.82& 2.75 & 2.31   \\
         L-DM &4.28&3.52& 3.16 \\
         LDGM &\textbf{4.30}&3.48&3.04 \\ 
        ASR & \textbf{4.30} & \textbf{3.66} & \textbf{3.39}\\ 
        \hline
    \end{tabular}
    \caption{Statistic Results(Average Score) of perceptual study}
    \vspace{-1em}
    \label{tab:my_label}
\end{table}

\subsection{Ablation Analysis}
Using large language model for layout generation can be challenged without proper input structure and overall pipeline. In this ablation study, we analyze the capability of relation matrix in allowing user interaction and generation reasonable results.

\paragraph{The Relation Matrix}
The latent relation matrix decoded by the graph neural network plays an indispensable role in the entire pipeline. Serving as a crucial input, the relation matrix enhances the large language model’s ability to accurately reconstruct graph relations—capabilities that do not naturally arise from the LLM’s original training phase.
In this part, we conducted supervised fine-tuning of the large language model both with and without the relation matrix to demonstrate the necessity and effectiveness of the graph representation decoder module. This comparison highlights how the relation matrix significantly improves the LLM's capacity to generate coherent and structurally consistent layouts, confirming its importance in the overall pipeline.

\begin{table}[]
    \centering
    \begin{tabular}{ccc}
    \hline
         & Max-IoU &Relation Error\\
         \hline
ASR w/o RM& 0.27& 0.61\\
ASR with RM &0.54 &0.30\\
\hline
    \end{tabular}
    \caption{Quantitative Results of ASR pipeline with or without Relation Matrix(RM)}
    \label{tab:woRM}
\end{table}
According to table \ref{tab:woRM}, using relation matrix can reduce the relation error significantly, contributing to final results such as Relation Error, Overlap, Max-IoU and FID. That is because large language model has strong ability for grounding, UI understanding and classification, which can generate accurate node attributes. However, the ability of generating relations between nodes are not learned. Thus, relation matrix can serve as a strong support for generating precise layout under a certain structure. 
Besides, the relation matrix also allow graph editing of user interaction while only using large language model cannot be able to do so. Moreover, unlike traditional graph-based generation, we take positional information together with semantic relations to generate reasonable and more complex results. As shown in the bottom of Figure\ref{fig:gallery}, the complex samples including more semantic information can be generated through our ASR.

\section{Conclusion}
Completing and generating UI layout designs has become increasingly important with the development of various terminal devices. Traditional vision-based layout generation methods often overlook structural information, while graph-based approaches lack the powerful capabilities of deep learning or large language models (LLMs). In our ASR, we combine graph representation with LLMs, preserving both the structural information of the graph and the predictive ability of deep language models. A graph representation learning module, incorporating a masking strategy and a relation matrix decoder block, was adopted to extract hierarchical representations from the given layout skeleton. We then applied a novel combination of graph features and relation matrices with an LLM to generate final layout designs. The reasonableness and diversity of these designs have been validated both quantitatively and qualitatively.

Our ASR introduces a new approach to aggregating graph features into LLMs for structural tasks. While this is primarily applied to layout generation, it is not limited to this task and can be adapted to a wide range of applications that involve structural information, such as scene and floorplan understanding and analysis. Moreover, we believe that with the support of LLMs, other graph-based methods, which have long faced challenges due to poor representation capabilities, will experience significant breakthroughs in the future.

\section*{Limitations}
\paragraph{Graph Conflicts}
The entire pipeline is not an end-to-end process, allowing for human-centered interaction. However, the unpredictability of human input can sometimes lead to conflicts between the previously generated relation matrix and manual graph modifications. In our pipeline, we prioritize human actions to resolve such conflicts when they are minimal. However, if there is a significant conflict with the existing graph, the system may struggle to generate a reasonable layout output.
\paragraph{Practical Limitations}
In our pipeline, we focus solely on generating UI layouts. However, in practical applications, UI designers still need to fill in the content for each component, which remains time-consuming. In the future, with the advanced capabilities of large language models (LLMs), it may be possible to generate the final UI directly, significantly reducing the time spent on manual adjustments.

\section*{Ethics Statement}
\paragraph{Human based User Study[D1 to D5]}
We conducted a crowdsourced user study using Amazon Mechanical Turk (AMT). The instruction provided to participants was: \textit{Score the following layout design based on your preference for its reasonableness. The score ranges from 1 to 5, where 1 indicates that the layout is not reasonable at all, and 5 indicates that it is perfectly reasonable.}

We paid participants according to the recommended pricing on the AMT system. All participants were formally informed before beginning their annotations that their responses would be used for academic research purposes.

\paragraph{AI utilization[E1]}
In our paper, we claim that we utilize AI(GPT) for polish our language. But we don't use AI for data processing and other operations that may conflict with academic ethics. 

\paragraph{Other Statements[B1 to B6]}
We cited all the used scientific artifacts among introduction, related work, experiments. The all codes and models we used are under the MIT license. We state that all the data and/or pretrained models are utilized compatible with their intention. No harmful attempts in our utilization. All the data we used are from open source dataset without any protective personal information and potentially offensive or risk data. The data we used include UI layout design image, English descriptions of each design image, json file indicating the structural information of each design. 
The pretrained models we used are open source models from HuggingFace, a LLM open model website.

\bibliography{custom}




\end{document}